\documentclass{article}
\usepackage[utf8]{inputenc}
\usepackage{graphicx}
\usepackage{url} 
\usepackage{amsmath}   
\usepackage{amssymb}   
\usepackage{authblk}
\usepackage{subcaption}
\usepackage{listings}
\usepackage{booktabs}
\usepackage[utf8]{inputenc}
\usepackage[T1]{fontenc}
\usepackage{listings}

\usepackage{algorithm}
\usepackage{algpseudocode}

\begin{document}

\title{SYNBUILD-3D: A large, multi-modal, and semantically rich synthetic dataset of 3D building models at Level of Detail 4}

\author[1]{Kevin Mayer\thanks{Corresponding author: kdmayer@stanford.edu}}
\author[2]{Alex Vesel}
\author[2]{Xinyi Zhao}
\author[1]{Martin Fischer}

\affil[1]{Department of Civil and Environmental Engineering, Stanford University, 473 Via Ortega, 94305, Stanford, USA}
\affil[2]{Department of Computer Science, Stanford University, 353 Jane Stanford Way, 94305, Stanford, USA}

\maketitle

\begin{abstract}
3D building models are critical for applications in architecture, energy simulation, and navigation. Yet, generating accurate and semantically rich 3D buildings automatically remains a major challenge due to the lack of large-scale annotated datasets in the public domain. Inspired by the success of synthetic data in computer vision, we introduce SYNBUILD-3D, a large, diverse, and multi-modal dataset of over 6.2 million synthetic 3D residential buildings at Level of Detail (LoD) 4. In the dataset, each building is represented through three distinct modalities: a semantically enriched 3D wireframe graph at LoD 4 (Modality I), the corresponding floor plan images (Modality II), and a LiDAR-like roof point cloud (Modality III). The semantic annotations for each building wireframe are derived from the corresponding floor plan images and include information on rooms, doors, and windows. Through its tri-modal nature, future work can use SYNBUILD-3D to develop novel generative AI algorithms that automate the creation of 3D building models at LoD 4, subject to predefined floor plan layouts and roof geometries, while enforcing semantic–geometric consistency. Dataset and code samples are publicly available at \url{https://github.com/kdmayer/SYNBUILD-3D}.
\end{abstract}

\begin{figure}[htbp]
    \centering
    \includegraphics[width=0.9\textwidth]{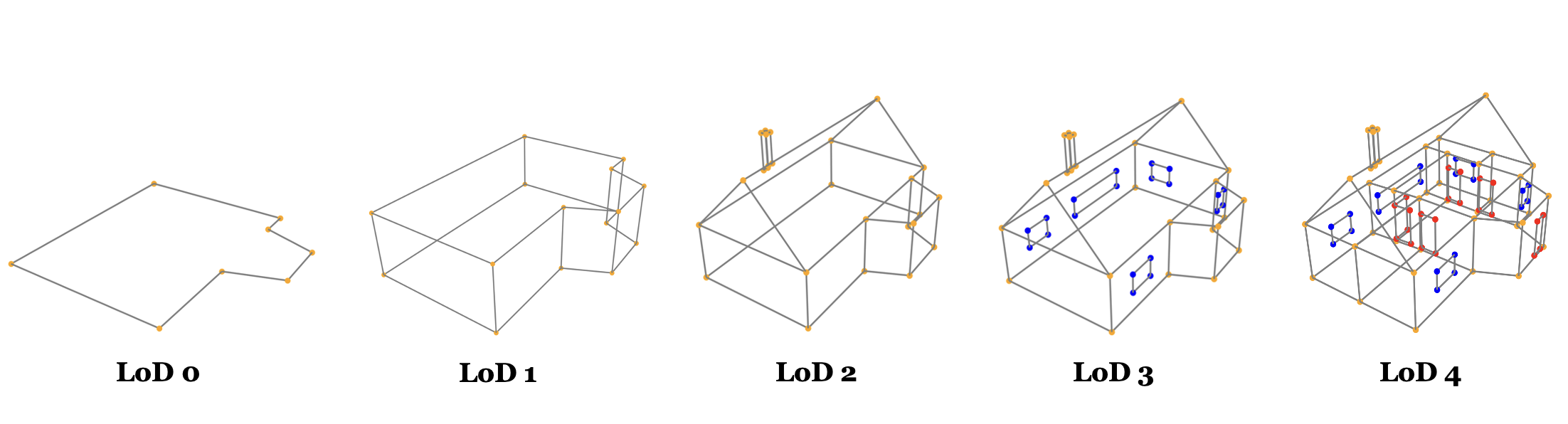}
    \caption{Overview comparing different Levels of Detail for 3D buildings.}
    \label{fig:lod_overview}
\end{figure}

\section{Background \& Summary}

3D building models serve as the foundation for numerous applications spanning architecture, design, energy simulation, and navigation. Despite significant advances in generative modeling across domains like images, text, and molecules, generating accurate and semantically rich 3D buildings remains a challenge in geometric deep learning. A primary obstacle to progress in this area is the scarcity of large-scale, semantically annotated, and diverse 3D building datasets in the public domain.

\subsection{Overview of Existing Datasets}

Existing 3D building datasets are commonly categorized according to their Level of Detail \cite{BILJECKI_LOD, CityGML_Overview}, a standardized classification that defines the geometric and semantic richness of 3D building representations as illustrated in Figure \ref{fig:lod_overview}. Below, we review representative datasets for each Level of Detail, with an emphasis on their type (real or synthetic), scale, and intended application area.

\subsubsection*{LoD 0–1: Footprint and Simple Block Models}

At the lowest Levels of Detail, LoD 0 and LoD 1 representations provide only building footprints or simple extruded block models without any roof geometry or semantic attributes. These models are often derived from GIS data \cite{OpenStreetMap}, overhead imagery \cite{GlobalMLBuildingFootprints}, or 2D cadastral maps \cite{NYS_Building_Footprints} and serve as coarse inputs for large-scale urban planning or navigation systems. While these datasets are typically real-world and available at a global scale, their lack of structural and semantic detail renders them insufficient for architectural design, generative modeling, simulation, and reconstruction tasks.

\subsubsection*{LoD 2: Roof-Enhanced Exterior Models}

LoD 2 adds simplified roof structures to extruded building footprints, offering more realistic silhouettes suitable for urban-scale energy simulations, urban planning, and 3D visualizations.

Prominent real-world datasets in this category include CityGML-based models from cities \cite{3D_Dresden, 3D_Prague, 3D_NYC, 3D_Zurich}, states \cite{3D_Bavaria, OpenNRW}, and countries \cite{3D_BAG, 3D_Japan, 3D_Poland}, collectively covering tens of millions of buildings \cite{CityGML_Overview}. 

Another notable dataset in this category is \cite{Building3D} which focuses specifically on roof wireframes. It comprises over 160,000 roof point clouds and corresponding roof wireframes extracted from airborne LiDAR, along with 47,000 simplified 3D building meshes and is particularly valuable for training and evaluating roof reconstruction algorithms \cite{PBWR_2024}.

Despite their large scale and practical value, LoD 2 datasets share a fundamental limitation: they do not include information about a building’s internal layout or its semantic components. While these datasets can be useful for coarse simulation and reconstruction tasks, they are insufficient for applications requiring a more detailed understanding in terms of geometry and semantics.

\subsubsection*{LoD 3: Detailed Exterior with Semantic Labels}

LoD 3 models capture detailed exteriors, e.g., windows, balconies, and façades, with semantic labels, enabling tasks like segmentation, façade parsing, and urban visualization.

A prominent example in this category is \cite{BuildingNetDataset} which provides labeled meshes and point clouds for over 2,000 synthetic buildings, surpassing the diversity and scale of real-world datasets like \cite{3D_Poznan, 3D_Ingolstadt}.

While useful for façade-aware generation and exterior simulations, LoD 3 models lack interior structure, limiting their use in holistic generative tasks and energy modeling.

\subsubsection*{LoD 4: Comprehensive Interior-Exterior Representations}

LoD 4 offers the highest detail, capturing full interior layouts, multistory structures, and the placement of elements like doors and windows, crucial for applications such as architectural modeling, indoor navigation, and energy simulation.

\begin{figure}[!htbp]
    \centering
    \includegraphics[width=0.7\textwidth]{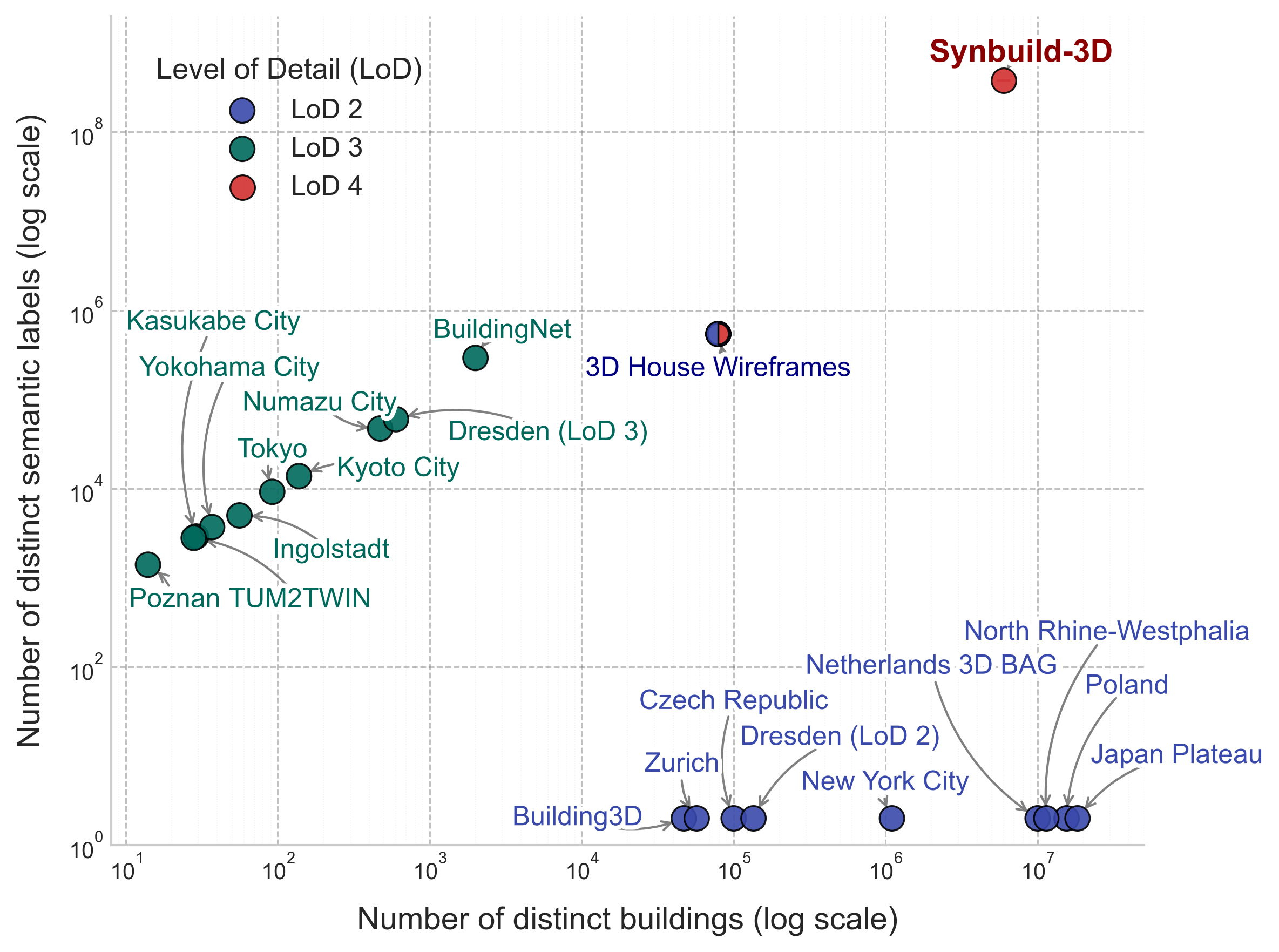}
    \caption{Comparison of public 3D building datasets by semantic diversity, scale, and detail. SYNBUILD-3D is over 100× more semantically diverse than prior datasets \cite{CityGML_Overview, Building3D, 3dHouseWireDataset, BuildingNetDataset}.}
    \label{fig:building_datasets_comparison}
\end{figure}

To the best of our knowledge, \cite{3dHouseWireDataset} is the only publicly available 3D building dataset that includes interior layouts, created by combining extruded floor plans from \cite{RPLAN} with heuristically generated roof structures. While the buildings in \cite{3dHouseWireDataset} are structurally close to LoD 4, their lack of semantic annotations for rooms, doors, and windows places them closer to LoD 2. In addition, the dataset is limited to single-story buildings with low roof diversity. Its moderate scale (approximately 78,000 buildings) further restricts its utility. Consequently, at this point, no existing 3D building dataset fully achieves LoD 4 in terms of structure and semantics.
\\
Figure \ref{fig:building_datasets_comparison} illustrates that existing datasets suffer from significant limitations: they offer either reasonable scale with minimal to no semantic detail (LoD 0-2) or richer semantics at a very limited scale (LoD 3-4). Moreover, most existing 3D building datasets focus solely on building exteriors. A notable exception is \cite{3dHouseWireDataset}, which leverages the floor plan dataset from \cite{RPLAN} to construct a collection of single-story residential buildings with interior room layouts. However, it lacks semantic annotations for rooms, doors, and windows. As a result, no existing 3D building dataset achieves LoD 4.

\subsection{Contribution}

To address these limitations and accelerate progress in 3D building generation, we introduce SYNBUILD-3D with the following contributions:

\begin{enumerate}

\item \textbf{Scale and Detail:} SYNBUILD-3D contains over 6.2 million synthetic residential buildings at LoD 4, each annotated with detailed room, door, and window geometries, providing over 390 million labels in total. Buildings span multiple floors with varying layouts, enabling realistic modeling of vertical architectural complexity.
\item \textbf{Unified Wireframes:} Unlike prior datasets that separate or omit structural components, SYNBUILD-3D provides integrated interior-exterior wireframes with consistent geometry and semantics, validated with input from building simulation experts.
\item \textbf{Multi-Modal Data:} Each building includes a semantically-enriched 3D wireframe, a roof point cloud, and floor plan images with segmentation masks, supporting conditional generation tasks across modalities.
\item \textbf{Open and Extensible:} We release the full dataset and generation pipeline to promote reproducibility and facilitate future extensions with new building types or features.

\end{enumerate}

\section{Methods}
\label{sec:methods}

\begin{figure}[!htbp]
    \centering
    \includegraphics[width=0.9\textwidth]{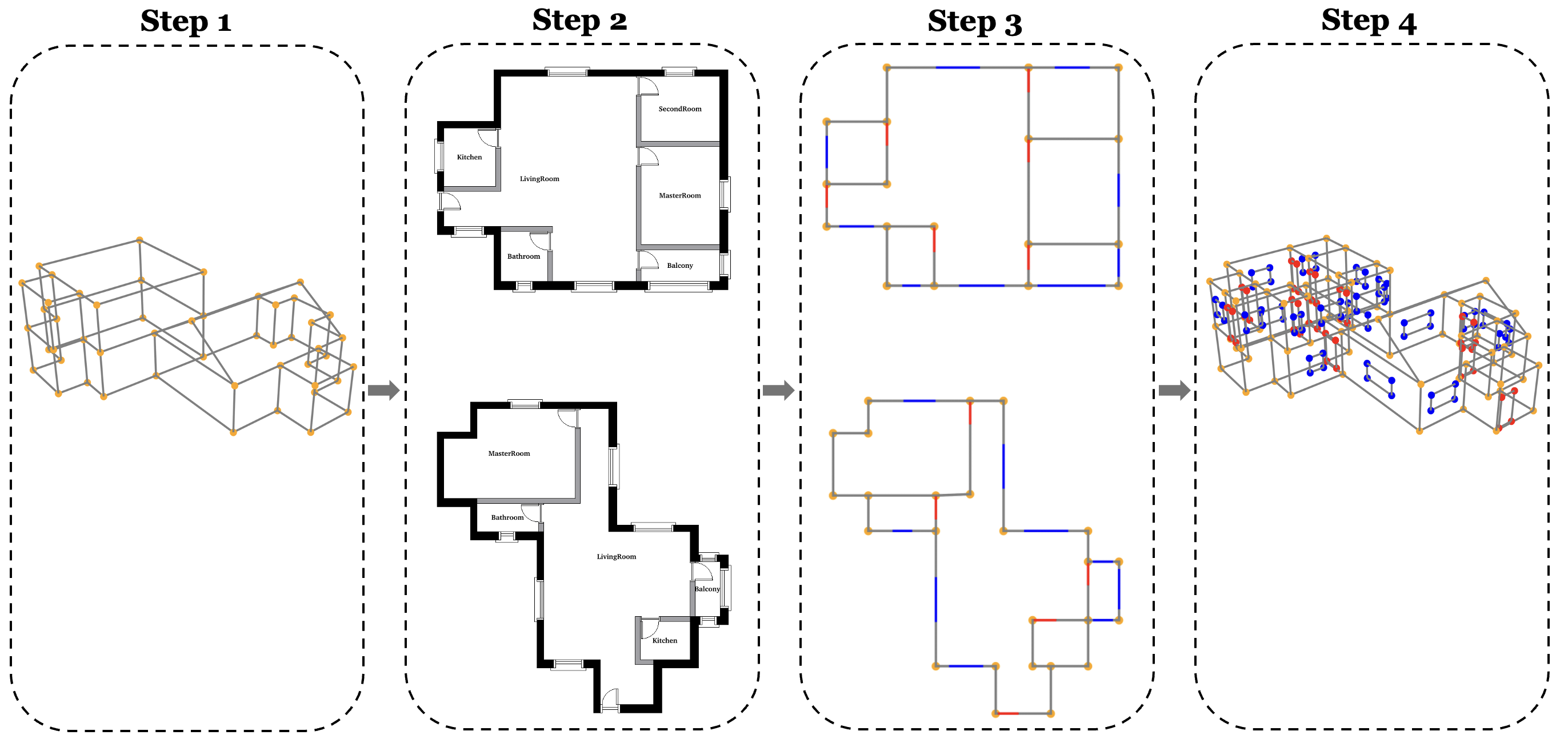}
    \caption{Overview of the SYNBUILD-3D pipeline. (1) A customized procedural generation engine first creates a randomized building exterior, extending the work of \cite{Random3dCity}. (2) For each floor, the corresponding footprint polygon is extracted and used to condition an AI-based floor plan generator \cite{RPLAN}, producing a floor-specific layout image. (3) The generated floor plans are then vectorized, and (4) the extruded floor volumes are stacked within the building hull.}
    \label{fig:pipeline_overview}
\end{figure}

To produce SYNBUILD-3D, we have developed a multi-stage pipeline for generating semantically enriched 3D wireframe models of residential buildings, including floor plan images with semantic segmentation masks, and a roof point cloud sampled at 50 points per square meter. The full process is illustrated in Figure~\ref{fig:pipeline_overview} and described below.

\subsection*{Step 1: Procedurally generate randomized building exteriors.}

We customize the procedural generation engine from \cite{Random3dCity} to create randomized building exteriors at LoD 3, which includes detailed roof superstructures and wall geometries. However, we discard all semantic annotations from the original output and retain only the geometric wireframe representation. The engine introduces diversity by randomizing parameters such as the footprint shape, number of stories, roof type (chosen from flat, gabled, hipped, pyramidal, or shed), and the presence of structural roof elements like dormers, chimneys, and roof windows. Since \cite{Random3dCity} can only generate buildings with rectangular footprints, we customize the existing engine by adding two randomized structural modifiers. The first enables the merging of two generated buildings into a single structure. The second introduces geometric complexity by appending up to four trapezoidal or rectangular extensions to a building footprint. As a result, both customizations enable the creation 3D building models with highly complex, non-rectangular footprint geometries. The first pipeline step thus produces a watertight 3D hull for each building, which is then decomposed into individual floor-level footprints for downstream processing.

\subsection*{Step 2: Generate footprint-conditioned floor plan images per floor.}

For each floor within a generated building hull, we extract the horizontal 2D footprint polygons to condition an AI-based floor plan generator, following \cite{RPLAN}. This means that Step 2 takes as input a set of floor-specific footprint boundaries and outputs corresponding floor plan images and segmentation masks. Since the floor plan generator's output depends on both the front door placement and the footprint area, we introduce layout diversity by randomly assigning a front door position along each footprint boundary. Figure \ref{fig:floorplan_generator_overview} in the appendix illustrates how different front door placements within a given footprint geometry can lead to distinct floor plan layouts.

\subsection*{Step 3: Vectorize floor plan information.}

In Step 3, we convert each floor plan image into a structured vector representation using its semantic segmentation mask, which labels every pixel by room type or structure. This mask guides the extraction of geometry and semantics, including room types and the positions and sizes of doors and windows. As existing floor plan vectorization algorithms assume Manhattan-world layouts \cite{R2V_Wu}, we developed a custom algorithm to handle non-Manhattan geometries. See Algorithm 1 in the appendix for details.

\subsection*{Step 4: Align, extrude, and stack vectorized floor volumes within building hull.} 

Lastly, we align the vectorized and semantically enriched floor plans with the original floor-specific footprint polygons and extrude them within the building volume produced in Step 1. During the alignment step, we represent each floor's footprint polygon and the corresponding floor plans as bitmaps such that pixels inside the building are set to one and pixels outside the building are set to zero. We then set up an optimization problem to determine the best parameters for translating ($t_x$, $t_y$) and scaling ($s_x$, $s_y$) the floor plan bitmap along the x- and y-axis on top of the floor's footprint bitmap. To do so, we compare the transformed floor plan bitmap to the original footprint bitmap across all pixels and minimize the following alignment loss:

\begin{align}
\label{eq:1}
\text{Loss}(t_x, t_y, s_x, s_y) &= 20 \cdot \text{Coverage} + \text{Overhang} \notag \\
&= \sum_{i,j} 20 \cdot \mathbf{1}_{{{Footprint Bitmap}_{ij}} > {{Floor Plan Bitmap(t_x, t_y, s_x, s_y)}_{ij}}} \notag \\
&\quad + \mathbf{1}_{{{Floor Plan Bitmap(t_x, t_y, s_x, s_y)}_{ij}} > {{Footprint Bitmap}_{ij}}}
\end{align}

\begin{figure}[!htbp]
    \centering
    \includegraphics[width=0.9\textwidth]{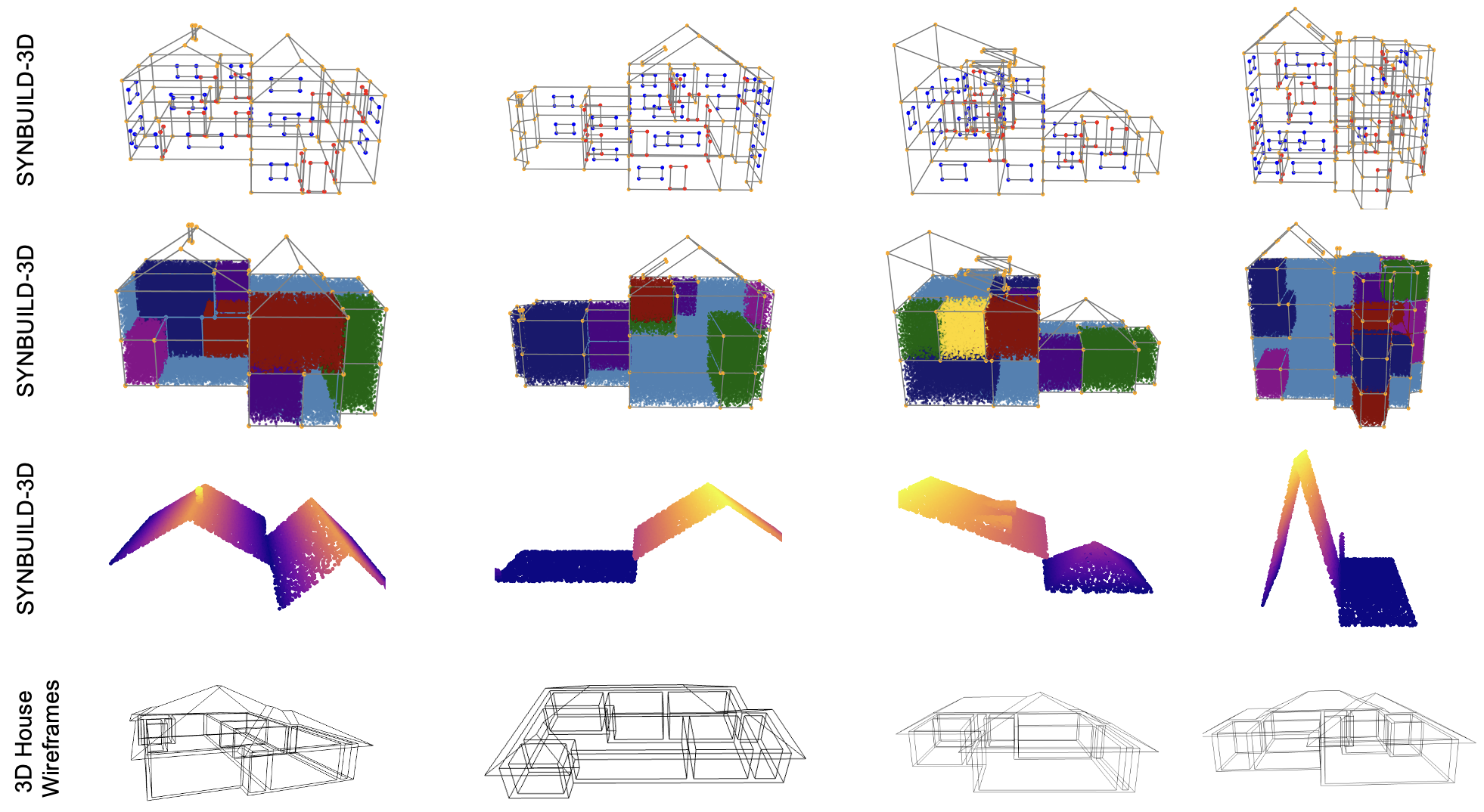}
    \caption{The top row shows four random SYNBUILD-3D wireframes, with windows in blue and doors in red. The second row adds color-coded room types derived from the building-specific floor plan images. The third row shows the corresponding roof point clouds. The bottom row displays wireframes from \cite{3dHouseWireDataset}, the most comparable dataset. Since room, door, or window annotations are not available, only the 3D structures are visualized.}
    \label{fig:dataset_visualization_overview}
\end{figure}

where \textit{Coverage} refers to the total number of pixels where the footprint bitmap extends beyond the transformed floor plan bitmap, while \textit{Overhang} represents the total number of pixels where the transformed floor plan bitmap exceeds the footprint bitmap area.
\\

Figure~\ref{fig:dataset_visualization_overview} presents four randomly selected buildings generated by our pipeline. To ensure high-quality building models, we generate multiple floor plan candidates per floor and retain only those that satisfy all quality criteria as described in Section~\ref{sec:Quality_Checks}. As a result, we often obtain more floor plans than the number of floors in a building. In such cases, we generate multiple distinct buildings by permuting the order in which floor plans are stacked. Conversely, when fewer floor plans are available, e.g., due to a non-satisfied quality criterion or an alignment failure, a valid floor plan may be reused once within a single building. 
\\

Dataset generation is distributed across 8 parallel pipelines, each running on a single AWS g5.2xlarge GPU instance for approximately 8 days.

\begin{figure}[!htbp]
    \centering
    \includegraphics[width=0.4\textwidth]{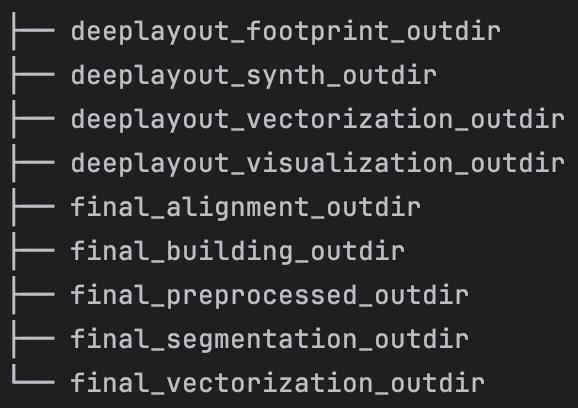}
    \caption{Dataset folder structure. The key directory to access the 3D wireframes is \textbf{final\_building\_outdir}. The floor plan images are in \textbf{deeplayout\_visualization\_outdir} and the segmentation masks are in \textbf{final\_segmentation\_outdir}.}
    \label{fig:folder_structure}
\end{figure}

\section{Data Records}
\label{sec:data_records}

The SYNBUILD-3D dataset can be downloaded from \url{https://purl.stanford.edu/kz908vb7844} and the code to produce the dataset is available at \url{https://github.com/kdmayer/SYNBUILD-3D}. 
\\

Since the dataset generation is distributed across 8 parallel pipelines, the full dataset is split across 8 separate .tar files and each .tar file is organized as illustrated in Figure \ref{fig:folder_structure}. The key directory to access the generated 3D wireframes is \textbf{final\_building\_outdir}. In this directory, users can find a set of folders and associated \textit{.JSON} files. An exemplary directory with two unique 3D building wireframes could look like this:

\begin{lstlisting}[basicstyle=\ttfamily]
final_building_outdir/
    building_001/
        final_building_abc.json
        final_building_xyz.json
\end{lstlisting}

In this example, we have two distinct 3D building wireframes, namely \textit{final\_building\_abc.json} and \textit{final\_building\_xyz.json}, both of which are generated by permuting valid floor plans within the exterior hull defined in \textit{building\_001}.
\\

Following guidance from domain experts in building modeling and simulation, each \textit{.JSON} file integrates interior and exterior geometry with semantic labels in a unified 3D wireframe. In particular, each \textit{.JSON} file includes 3D points and adjacency matrices for the building structure, windows, doors, and roof, along with high-resolution roof point clouds at 50 points per square meter, floor plan images with segmentation masks, and per-floor 3D wireframes:

\begin{figure}[!htbp]
    \centering
    \includegraphics[width=0.8\textwidth]{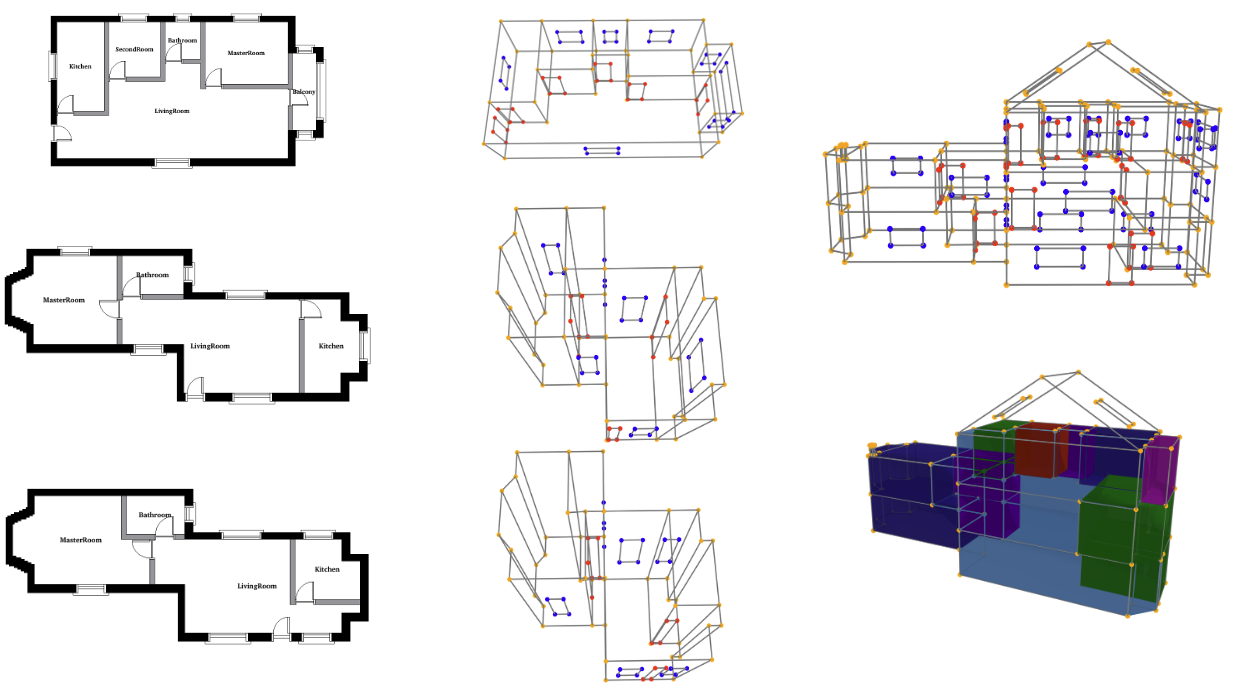}
    \caption{A three-story building with floor plan images (left), vectorized floor units including door and window semantics (middle), and the final 3D building wireframe with door and window semantics (right top) and room type information (right bottom).}
    \label{fig:dataset_visualization}
\end{figure}

\begin{itemize}
    \item \textit{final\_building\_points}: A set of N 3D points which define the structure of a building's 3D wireframe.
    \item \textit{final\_building\_adj}: The NxN adjacency matrix which defines how the N points in \textit{final\_building\_points} are connected to each other to create the structure of the 3D building wireframe.
    \item \textit{final\_window\_points}: A set of M 3D points which define the floor plan-derived window positions. 
    \item \textit{final\_window\_adj}: The MxM adjacency matrix which defines how the M points in \textit{final\_window\_points} are connected to form the windows.
    \item \textit{final\_door\_points}: A set of K 3D points which define the floor plan-derived door positions. 
    \item \textit{final\_door\_adj}: The KxK adjacency matrix which defines how the K points in \textit{final\_door\_points} are connected to form the doors.
    \item \textit{final\_roof\_points}: A set of L 3D points which define the roof geometry.
    \item \textit{final\_roof\_adj}: The LxL adjacency matrix which defines how the L points in \textit{final\_roof\_points} are connected to form the roof geometry.
    \item \textit{final\_room\_type\_dict}: A mapping of room type IDs, as defined in Table \ref{tab:room_type_ids} in the appendix, to 3D point indices from \textit{final\_building\_points}.
    \item \textit{unit\_dict\_list}: A hash map which defines the 3D wireframes for each individual floor unit, including windows, doors, and room boundaries.
    \item \textit{floorplan\_ID\_list}: A list of ordered floor plan IDs associated with the floor plan images in \textbf{deeplayout\_visualization\_outdir}. The first floor plan ID refers to the floor plan image associated with the lowest floor and the last floor plan ID refers to the floor plan image associated with the top floor.
    \item \textit{sampled\_roof\_points\_list}: A LiDAR-like point cloud sampled from the surface of the roof geometry.
\end{itemize}

\section{Technical Validation}
\label{sec:Quality_Checks}

\subsection{Automated Quality Checks}
\label{sec:automated_quality_checks}

To ensure the semantic and structural integrity of our dataset, we implement a series of automated quality control measures during generation:

\begin{itemize}
    \item Minimum Room Count: Every floor plan must contain at least three distinct rooms, ensuring sufficient spatial diversity and functional utility.
    \item Semantic Coverage: The entire building footprint must be fully annotated, i.e., no empty or unassigned regions within the footprint are allowed. Each area is labeled as a room, wall, door, or window.
	\item Room Enclosure Constraint: Every room must be completely surrounded by structural elements, ensuring spatial closure and preventing topological errors.
	\item Door–Room Consistency: Each room must be accessible via at least one door, and the number of doors must equal the number of rooms, enforcing functional access and realism.
    \item Unique Vectorized Nodes: All nodes are unique across the building. Repeated points across different floors, e.g., on the footprint boundary, are merged into a single shared node. This ensures a clean and consistent topological graph and prevents redundancy in the vector representation.
\end{itemize}

\subsection{Dataset Analyses}

Using our pipeline from Section \ref{sec:methods}, we generate 16,000 building exteriors and approximately 240,000 candidate floor plans. After quality filtering (Section \ref{sec:automated_quality_checks}), 8,698 exteriors (54.36\%) and 81,223 floor plans (33.84\%) are retained. Valid floor plans are then permuted across floors to synthesize 6.28 million distinct 3D buildings. If a building lacks sufficient unique floor plans for all floors, a valid plan may be reused once. When the top floor has a different footprint, only the lower floor plans are permuted, while a valid top floor plan is randomly selected.

\begin{figure*}[t]
  \centering
  \begin{subfigure}[t]{0.4\textwidth}
    \includegraphics[width=\linewidth]{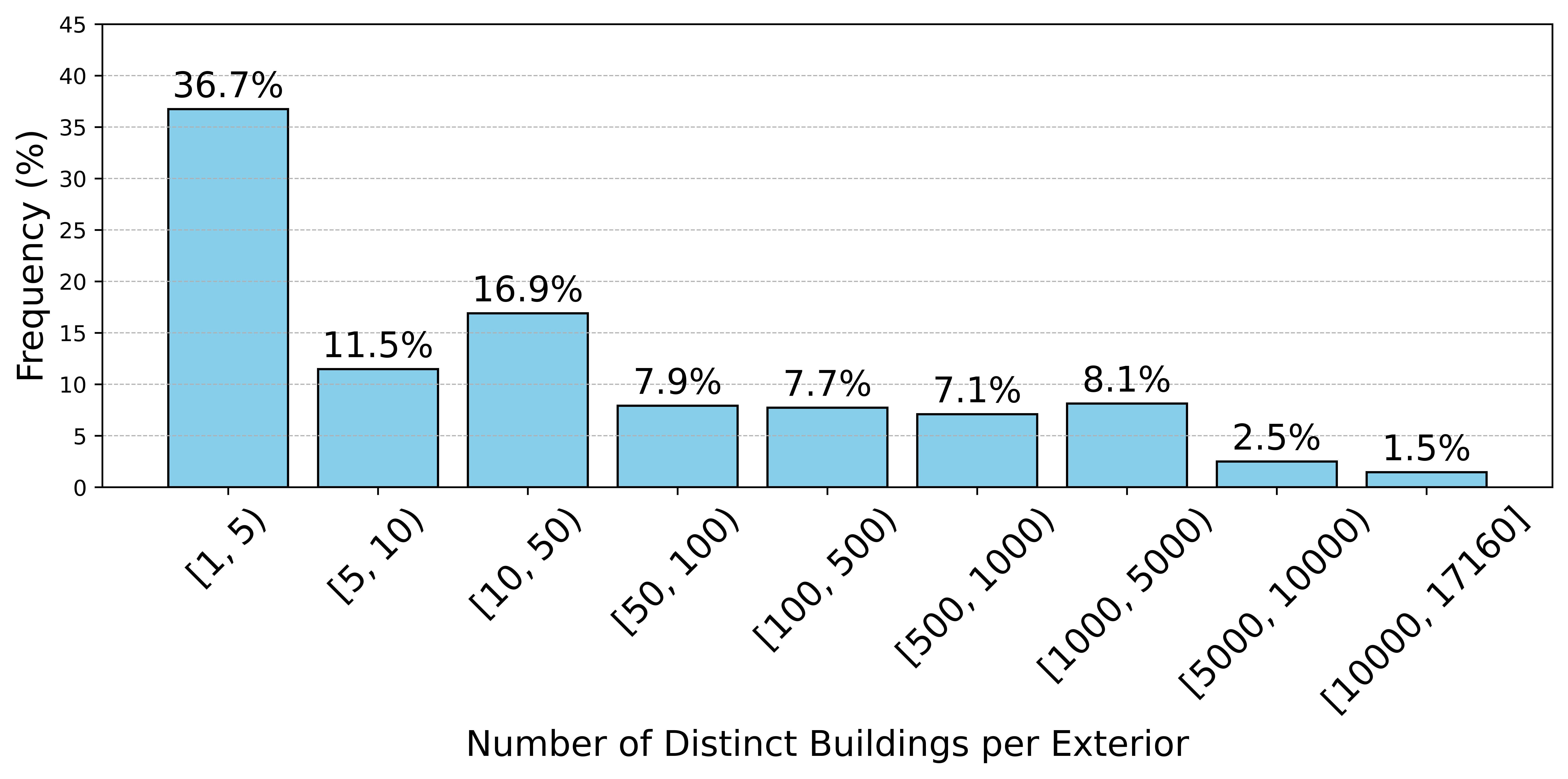}
    \caption{}
    \label{fig:subfig_a}
  \end{subfigure}
  \hfill
  \begin{subfigure}[t]{0.4\textwidth}
    \includegraphics[width=\linewidth]{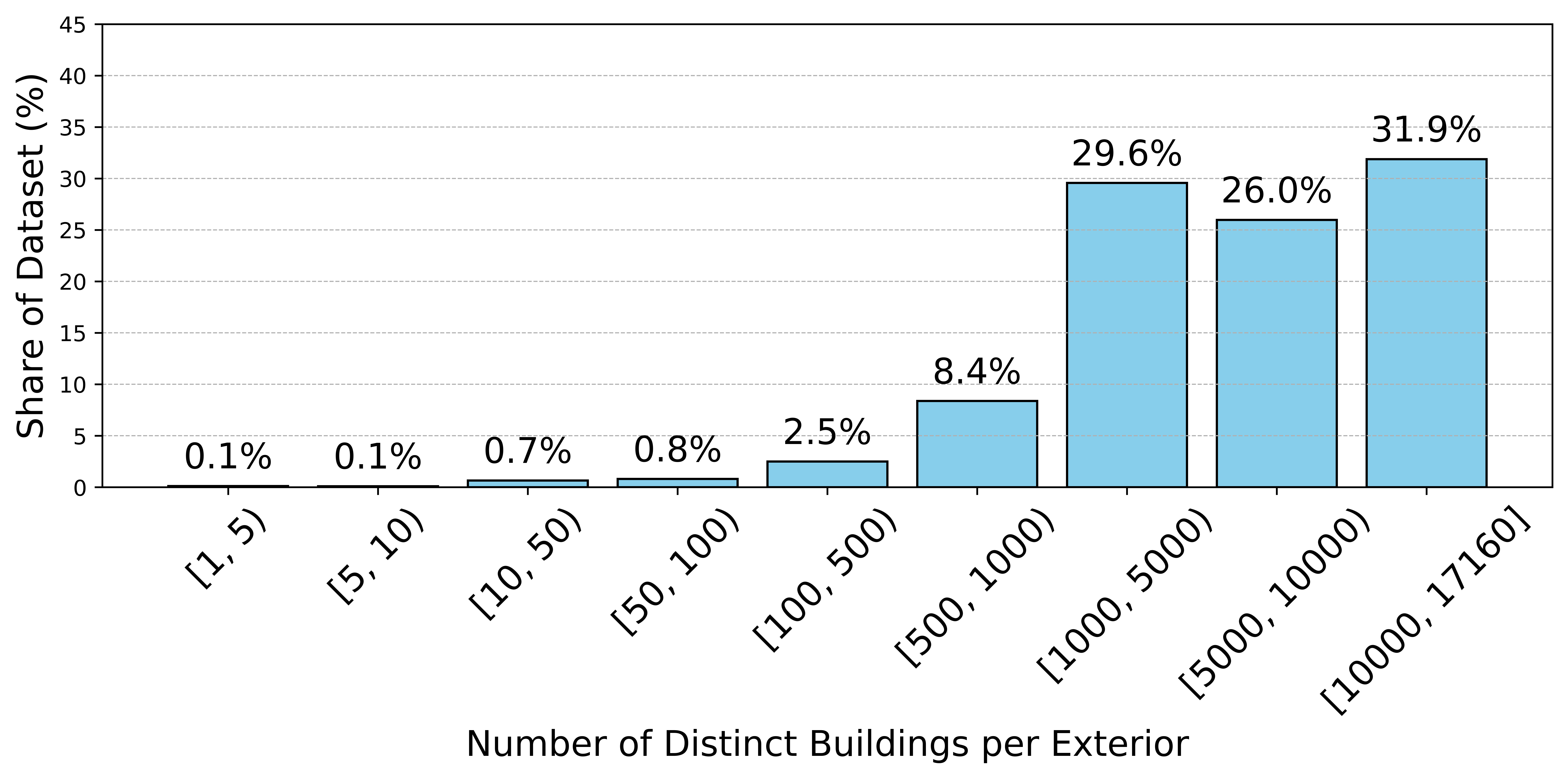}
    \caption{}
    \label{fig:subfig_b}
  \end{subfigure}
  \hfill
  \begin{subfigure}[t]{0.08\textwidth}
    \raisebox{10pt}{\includegraphics[width=\linewidth]{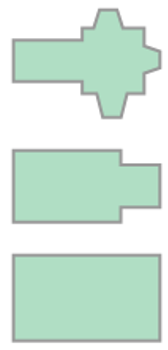}}
    \caption{}
    \label{fig:subfig_c}
  \end{subfigure}
  \caption{(a) Distribution of the number of distinct buildings per exterior. (b) Proportion of dataset by number of distinct buildings per exterior. (c) Selected building footprints for which more than 10,000 distinct buildings have been created via floor plan permutations.}
  \label{fig:dataset_statistics}
\end{figure*}

Figure \ref{fig:dataset_statistics} illustrates the role floor plan permutations play in scaling up the size of our dataset. Based on a random sample of 479 building exteriors and the associated 310,511 distinct buildings, Figure \ref{fig:subfig_a} describes the distribution of the number of distinct buildings per exterior. In other words, 36.7\% of building exteriors are associated with 1 to 4 distinct 3D building models, while 1.5\% of exteriors amass between 10,000 to 17,160 distinct building models. As an example, assuming a building exterior with 4 floors is associated with 12 unique floor plans, it is possible to create ${}_{n}P_{r} = {}_{12}P_{4} = \frac{12!}{(12 - 4)!} = 11,880$ distinct 3D buildings. Figure \ref{fig:subfig_b} shows the proportion of the dataset by the number of distinct buildings per exterior. As such, we can see that 87.5\% of the dataset comes from building exteriors that have between 1,000 and 17,160 distinct floor plan permutations. As a result, we can conclude that roughly 12.1\% (1,052) of building exteriors are associated with 87.5\% (5,497,786) of the distinct buildings in the dataset. While this means that there is an imbalance in the dataset towards a subset of the generated building exteriors, the respective exteriors can nonetheless be quite complex, as illustrated in the top element of Figure \ref{fig:subfig_c} which displays a building footprint with three trapezoidal extensions for which more than 10,000 distinct buildings are created. 
Even within each building exterior, the randomness of the floor plan generator and the presence of multiple footprint shapes across floors lead to significant variance as illustrated in Figure 7 in the appendix. In general, buildings with more floors benefit disproportionately from floor plan permutation, resulting in greater number of buildings with multistory layout variations.

\subsection{Manual Quality Assessment}

To verify the quality of the final dataset, we conduct a manual inspection on 250 randomly sampled building wireframes, including 993 floor plans, 6,461 windows, 4,593 doors, 20,946 wall segments. From our manual validation, we find that 10.8\% of buildings exhibit a \textit{single} minor error. We thereby define a minor error as a missing interior wall segment or a missing door or window element. In addition, we find that 1.6\% of buildings exhibit major errors such as an unreachable room, a missing exterior wall segment, or a combination of two or more minor errors. On the floor plan level, 2.8\% exhibit minor errors and 0.4\% show major errors. 
\\
Element-wise accuracy is high: 99.95\% of walls, 100\% of windows, and 99.78\% of doors are correct. Notably, 32.8\% of floor plans and 36\% of wireframes feature complex, non-orthogonal geometries across up to four floors, significantly extending existing datasets' semantic richness and geometric complexity. 

\subsection{Building-level Statistics}

Sampling 10,000 random buildings and computing building-level statistics in terms of the generated wireframe graphs, we can see that each wireframe graph has between 84--512 nodes (median 308, std. 50.92), 110--705 edges (median 427, std. 72.24), and 1--4 floors (median 4, std. 0.23). The random sample contains 185,137 distinct rooms of which 21.37\% are living rooms, 19.45\% are kitchens, 18.31\% are bathrooms, 15.88\% are master bedrooms, 14.64\% are secondary bedrooms, 9.56\% are balconies, and 0.79\% are study rooms. Across the random sample, we observe 629,613 labeled elements, i.e., rooms, doors, and windows with unique 3D positions, which we extrapolate to around 395 million semantic labels across the full dataset.

\section{Data Availability}

The SYNBUILD-3D dataset can be downloaded from \url{https://purl.stanford.edu/kz908vb7844}.

\section{Code Availability}

Code to visualize the 3D building wireframes is available at \url{https://github.com/kdmayer/SYNBUILD-3D}. The repository includes a detailed description of software and python packages used, as well as their versions. We will release the full pipeline for generating the 3D wireframes after the peer-review process.

\section{Credits and acknowledgements}

\noindent \textit{Kevin Mayer}: Conceptualization, Methodology, Software, Validation, Formal analysis, Investigation, Visualization, Data curation, Writing – original draft, Writing – review \& editing, Supervision, Funding acquisition.\\
\textit{Alex Vesel}: Software, Methodology, Investigation. \\
\textit{Xinyi Zhao}: Software.\\
\textit{Martin Fischer}: Conceptualization, Supervision, Writing – original draft, Project administration, Funding acquisition, Resources.\\
\\
\noindent This research was supported by Stanford’s Bits\&Watts initiative in collaboration with E.ON SE and through CIFE Seed funds sponsored by Goldbeck GmbH. Kevin Mayer would like to express his gratitude to Stefan Padberg, Dr. Munib Amin, Tim Schoenheit, Stefan Gehder, Dr. Alejandro Newell, and Chris Agia. Kevin Mayer would like to express his gratitude to Isabel Larus for her support throughout the research project. Kevin Mayer would like to thank the Satre family for sponsoring his Stanford Interdisciplinary Graduate Fellowship.
\\

\noindent The author(s) declare no competing interests.

\section{Appendix}

\subsection{Floor Plan Vectorization Algorithm}

\begin{algorithm}
\caption{Floor Plan Vectorization Algorithm}
\label{alg:vectorization_algorithm}
\begin{algorithmic}[1]
\Procedure{Vectorizer}{bitmap}
  \State \textit{Input:} Binary floor plan bitmap ($1$: structure, $0$: empty)
  \State \textit{Output:} Vectorized floor plan graph (nodes and adjacency matrix)

  \State \textbf{Initialize graph}
  \State \hspace{1em} Place nodes randomly on pixels where bitmap = 1
  \State \hspace{1em} Connect neighboring node pairs if pixels along line segment are 1

  \For{each refinement iteration}
    \State Identify and preserve structural anchor nodes (e.g., wall corners)
    \State Merge non-anchor nodes with closest anchor node
    \State Align node positions to respect wall directions
  \EndFor

  \State Ensure nearest neighbor connectivity in each direction
  \State \Return Vectorized floor plan graph
\EndProcedure
\end{algorithmic}
\end{algorithm}

\subsection{Floor Plan Generation Examples}

\begin{figure}[H]
    \centering
    \includegraphics[width=0.7\textwidth]{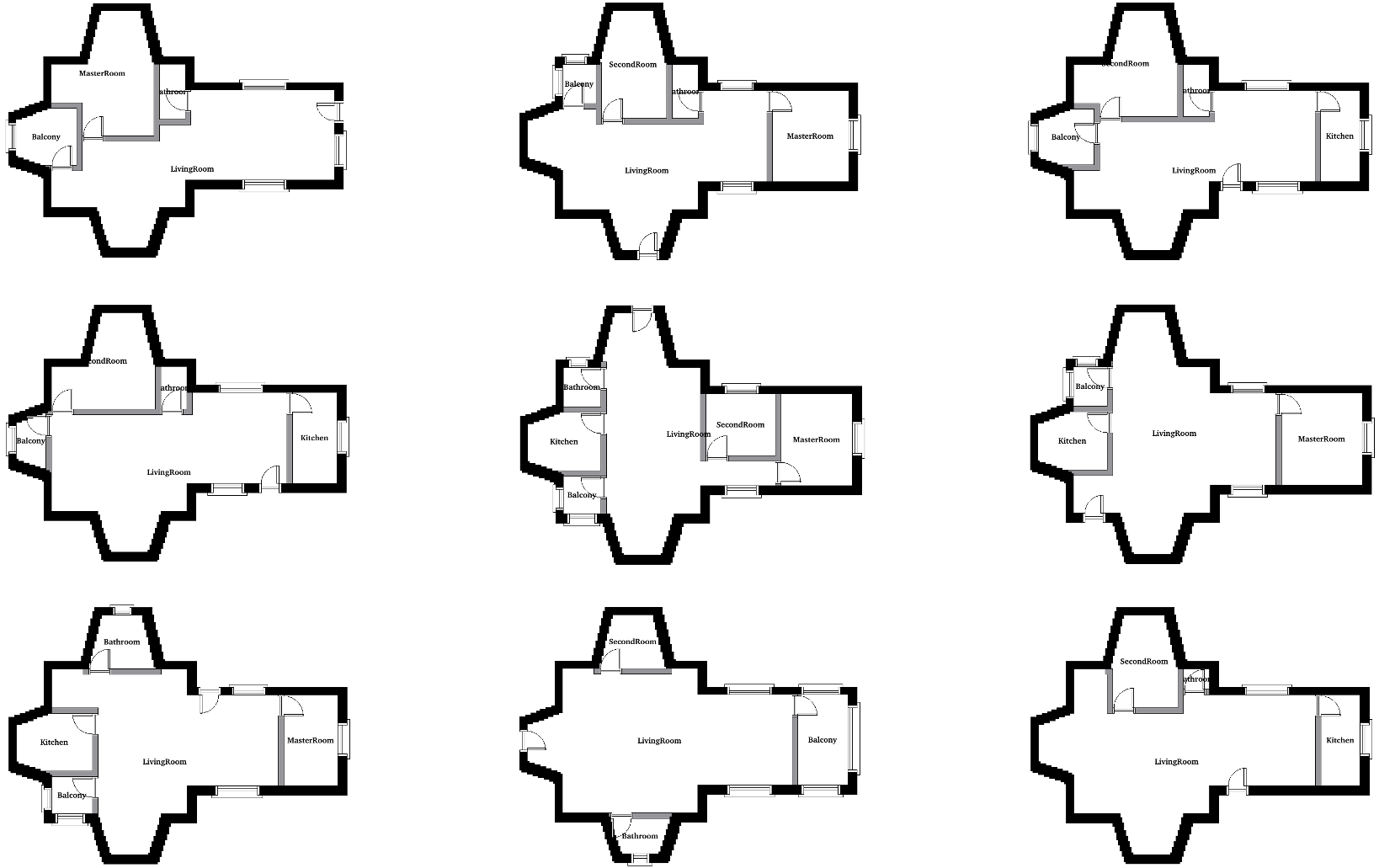}
    \caption{We use \cite{RPLAN} to generate floor plans based on building footprint boundaries. This figure illustrates nine distinct floor plans produced for the same footprint. In general, variations arise from both the footprint geometry and the randomized placement of the front door.}
    \label{fig:floorplan_generator_overview}
\end{figure}

\subsection{Room Type to Label Mapping}

Following \cite{RPLAN}, Table \ref{tab:room_type_ids} illustrates the mapping of room types and elements to integer IDs.

\begin{table}[!h]
\centering
\caption{Room and Element Labels}
\begin{tabular}{ll}
\toprule
\textbf{Name} & \textbf{ID} \\
\midrule
Living Room     & 1  \\
Master Room     & 2  \\
Kitchen        & 3  \\
Bathroom       & 4  \\
Dining Room     & 5  \\
Child Room      & 6  \\
Study Room      & 7  \\
Second Room     & 8  \\
Guest Room      & 9  \\
Balcony        & 10 \\
Entrance       & 11 \\
Storage        & 12 \\
Wall-in        & 13 \\
External       & 14 \\
Exterior Wall   & 15 \\
Interior Wall   & 16 \\
Front Door      & 17 \\
Interior Door   & 18 \\
Open Wall       & 19 \\
Window         & 20 \\
Balcony Door    & 21 \\
\bottomrule
\label{tab:room_type_ids}
\end{tabular}
\end{table}

\newpage

\bibliographystyle{unsrt}
\bibliography{main}

\begin{thebibliography}{10}

\bibitem{BILJECKI_LOD}
Filip Biljecki, Hugo Ledoux, and Jantien Stoter.
\newblock An improved lod specification for 3d building models.
\newblock {\em Computers, Environment and Urban Systems}, 59:25--37, 2016.

\bibitem{CityGML_Overview}
O.~Wysocki, B.~Schwab, C.~Beil, C.~Holst, and T.~H. Kolbe.
\newblock Reviewing open data semantic 3d city models to develop novel 3d reconstruction methods.
\newblock {\em The International Archives of the Photogrammetry, Remote Sensing and Spatial Information Sciences}, XLVIII-4-2024:493--500, 2024.

\bibitem{OpenStreetMap}
{OpenStreetMap contributors}.
\newblock {Planet dump retrieved from https://planet.osm.org }.
\newblock \url{ https://www.openstreetmap.org }, 2017.

\bibitem{GlobalMLBuildingFootprints}
{Microsoft}.
\newblock {Worldwide building footprints derived from satellite imagery}.
\newblock \url{ https://github.com/microsoft/GlobalMLBuildingFootprints }, 2024.

\bibitem{NYS_Building_Footprints}
{State of New York}.
\newblock {NYS Building Footprints}.
\newblock \url{ https://data.gis.ny.gov/maps/a6bbc64e38f04c1c9dfa3c2399f536c4/about }, 2025.

\bibitem{3D_Dresden}
{State of Saxony}.
\newblock {3D Buildings Dresden }.
\newblock \url{ https://www.geodaten.sachsen.de/downloadbereich-digitale-3d-stadtmodelle-4875.html }, 2021.

\bibitem{3D_Prague}
The~Institute of~Planning and Development.
\newblock {3D Buildings}.
\newblock \url{ https://geoportalpraha.cz/en/data-and-services/articles-and-projects/3d-model}, 2024.

\bibitem{3D_NYC}
{New York City Department of City Planning}.
\newblock {NYC 3D Model}.
\newblock \url{ https://www.nyc.gov/content/planning/pages/resources/datasets/nyc-3d-model }, 2014.

\bibitem{3D_Zurich}
{City of Zurich}.
\newblock {Zurich 3D Buildings}.
\newblock \url{https://3d.stzh.ch/appl/3d/zuerichvirtuell/}, 2025.

\bibitem{3D_Bavaria}
State of~Bavaria.
\newblock {3D-Gebäudemodell}.
\newblock \url{ https://geodaten.bayern.de/opengeodata/}, 2025.

\bibitem{OpenNRW}
{State of North Rhine-Westphalia}.
\newblock {LoD2 Gebaeudemodelle}.
\newblock \url{ https://www.opengeodata.nrw.de/produkte/geobasis/3dg/lod2_gml/ }, 2017.

\bibitem{3D_BAG}
Balázs Dukai, Ravi~Y. Peters, Joris N.~H. van Liempt, Jantien~E. Stoter, Stefan Vitalis, and Tianyu Wu.
\newblock {3D BAG}.
\newblock \url{ https://3dbag.nl}, 2021.

\bibitem{3D_Japan}
MLIT Japan.
\newblock {Project Plateau}.
\newblock \url{ https://plateau.takram.com/}, 2023.

\bibitem{3D_Poland}
Poland's Central~Office for Geodesy and Cartography.
\newblock {3D Building Models}.
\newblock \url{ https://mapy.geoportal.gov.pl/imap/Imgp_2.html gpmap=imap3d}, 2022.

\bibitem{Building3D}
R.~Wang, S.~Huang, and H.~Yang.
\newblock Building3d: An urban-scale dataset and benchmarks for learning roof structures from point clouds.
\newblock In {\em 2023 IEEE/CVF International Conference on Computer Vision (ICCV)}, pages 20019--20029, Los Alamitos, CA, USA, oct 2023. IEEE Computer Society.

\bibitem{PBWR_2024}
Shangfeng Huang, Ruisheng Wang, Bo~Guo, and Hongxin Yang.
\newblock Pbwr: Parametric-building-wireframe reconstruction from aerial lidar point clouds.
\newblock In {\em Proceedings of the IEEE/CVF Conference on Computer Vision and Pattern Recognition (CVPR)}, pages 27778--27787, June 2024.

\bibitem{BuildingNetDataset}
Pratheba Selvaraju, Mohamed Nabail, Marios Loizou, Maria Maslioukova, Melinos Averkiou, Andreas Andreou, Siddhartha Chaudhuri, and Evangelos Kalogerakis.
\newblock Buildingnet: Learning to label 3d buildings.
\newblock In {\em Proceedings of the IEEE/CVF International Conference on Computer Vision (ICCV)}, pages 10397--10407, October 2021.

\bibitem{3D_Poznan}
City of~Poznan.
\newblock {3D Buildings LoD 3}.
\newblock \url{ https://sipgeoportal.geopoz.poznan.pl/}, 2024.

\bibitem{3D_Ingolstadt}
Savenow.
\newblock {LoD 3 Road Space Models}.
\newblock \url{ https://github.com/savenow/lod3-road-space-models/}, 2024.

\bibitem{3dHouseWireDataset}
Xueqi Ma, Yilin Liu, Wenjun Zhou, Ruowei Wang, and Hui Huang.
\newblock Generating 3d house wireframes with semantics.
\newblock In {\em ECCV}, 2024.

\bibitem{RPLAN}
Wenming Wu, Xiao-Ming Fu, Rui Tang, Yuhan Wang, Yu-Hao Qi, and Ligang Liu.
\newblock Data-driven interior plan generation for residential buildings.
\newblock {\em ACM Trans. Graph.}, 38(6), November 2019.

\bibitem{Random3dCity}
Filip Biljecki, Hugo Ledoux, and Jantien Stoter.
\newblock {Generation of multi-LOD 3D city models in CityGML with the procedural modelling engine Random3Dcity}.
\newblock {\em ISPRS Ann. Photogramm. Remote Sens. Spatial Inf. Sci.}, pages 51--59, 2016.

\bibitem{R2V_Wu}
Chen Liu, Jiajun Wu, Pushmeet Kohli, and Yasutaka Furukawa.
\newblock Raster-to-vector: Revisiting floorplan transformation.
\newblock In {\em Proceedings of the IEEE International Conference on Computer Vision (ICCV)}, Oct 2017.

\end{thebibliography}

\end{document}